\documentclass[a4paper]{article}

\usepackage{INTERSPEECH2022}
\usepackage{amsmath,amssymb,graphicx,amsfonts,cite,tabularx,multirow,xcolor,url,tipa,breakurl,bm}
\usepackage[htt]{hyphenat}

\usepackage{xcolor}
\usepackage{amsmath}
\usepackage{makecell}

\title{Deliberation Model for On-Device Spoken Language Understanding}
\name{
    Duc Le$^*$,
    Akshat Shrivastava$^*$,
    Paden Tomasello,
    Suyoun Kim,\\
    Aleksandr Livshits,
    Ozlem Kalinli,
    Michael L. Seltzer\thanks{$^*$Equal contribution.}
}
\address{Meta, USA}
\email{\{duchoangle,akshats\}@fb.com}

\begin{document}
\ninept

\maketitle
\begin{abstract}
We propose a novel deliberation-based approach to end-to-end (E2E) spoken language understanding (SLU), where a streaming automatic speech recognition (ASR) model produces the first-pass hypothesis and a second-pass natural language understanding (NLU) component generates the semantic parse by conditioning on both ASR's text and audio embeddings. By formulating E2E SLU as a generalized decoder, our system is able to support complex compositional semantic structures. Furthermore, the sharing of parameters between ASR and NLU makes the system especially suitable for resource-constrained (on-device) environments; our proposed approach consistently outperforms strong pipeline NLU baselines by 0.60\% to 0.65\% on the spoken version of the TOPv2 dataset (STOP). We demonstrate that the fusion of text and audio features, coupled with the system's ability to rewrite the first-pass hypothesis, makes our approach more robust to ASR errors. Finally, we show that our approach can significantly reduce the degradation when moving from natural speech to synthetic speech training, but more work is required to make text-to-speech (TTS) a viable solution for scaling up E2E SLU.
\end{abstract}
\noindent\textbf{Index Terms}: spoken language understanding, end-to-end training, deliberation method, on-device processing

\vspace{-0.5em}
\section{Introduction}
\label{sec:intro}

Traditional Spoken Language Understanding (SLU) systems typically consist of two separate components. The automatic speech recognition (ASR) component generates transcription from audio, while the natural language understanding (NLU) component generates semantic information from the ASR hypothesis. This \textbf{pipeline} approach allows the two components to be developed separately, with ASR trained on labeled audio data and NLU trained on text-only data, leading to faster iteration speed and easier maintainability. On the other hand, this approach also comes with several limitations, such as being prone to error propagation from ASR to NLU, lack of acoustic information which limits NLU's accuracy, and lack of parameter sharing which makes it difficult to bring SLU on-device.

\textbf{End-to-end} (E2E) SLU systems \cite{Haghani18SLU,serdyuk2018endtoend,Potdar2021ASE, Radfar2021FANSFA,Wang2021Speech2SlotAE,2020NeuralInterface,raju2021end} attempt to overcome these limitations by going directly from audio to semantics. While this direction has shown promise, current E2E systems face three key challenges. Firstly, E2E systems are often treated as a black box of audio to semantics without the ability to output transcripts~\cite{Haghani18SLU,serdyuk2018endtoend,Radfar2021FANSFA,Wang2021Speech2SlotAE,Potdar2021ASE}. In practice, transcript generation is a requirement for many speech-based applications; in addition, having access to the transcripts is beneficial for debugging and understanding the system's behavior. Secondly, E2E SLU often targets domain/intent prediction \cite{Haghani18SLU,serdyuk2018endtoend} or slot tagging \cite{Radfar2021FANSFA,Wang2021Speech2SlotAE,Potdar2021ASE}, but does not appear to solve complex understanding use cases (e.g., composition) that text based systems are capable of~\cite{decoupled, gupta2018semantic, rongali2020don}. Thirdly, certain E2E SLU solutions have been proposed that combine text and audio data via a neural interface to produce the semantic parse and thus retain the ability to output transcripts~\cite{2020NeuralInterface,raju2021end}. However, these systems are understudied from an efficiency, robustness, and scalability perspective. It remains unclear how suitable they are in resource-constrained (on-device) environments, how robust they are against speech variation and ASR errors, and how scalable they are to new domains given the reliance on audio training data.

In this work, we propose a novel deliberation-based approach to E2E SLU to address these challenges. Our method is inspired by deliberation-based two-pass E2E ASR~\cite{Hu20deliberation}, where a neural correction model is conditioned on both the acoustics and the first-pass hypothesis (often generated by a streaming ASR model) to produce the final transcript. Unlike in deliberation-based ASR, our second-pass model generates the semantic parse instead of a corrected version of the first-pass hypothesis. Our system naturally retains the ability to output transcripts via the first-pass streaming ASR model, thus addressing the first challenge. By formulating E2E SLU as a generalized decoder instead of a collection of classification models, we enable the system to support complex compositional semantic structures, addressing the second challenge. Our in-depth analysis reveals that the proposed approach consistently outperforms strong pipeline NLU baselines by \textbf{0.60\%} to \textbf{0.65\%} in accuracy under two operating points and is especially suitable for resource-constrained on-device environments. We demonstrate that our model's implicit ability to rewrite the first-pass hypothesis makes it more robust against ASR errors compared to traditional pipeline NLU systems. Finally, we show that the system's multimodal nature significantly lessens the degradation when shifting from natural speech to synthetic speech training data. However, more work is required to further close the gap and make text-to-speech (TTS) a viable solution for scaling up E2E SLU without collecting natural speech.

\vspace{-0.5em}
\section{Deliberation-Based End-to-End SLU}
\label{sec:method}

\begin{table*}
\centering
\caption{Example NLU annotations for the music and navigation domains in TOPv2 (some ontology tokens are shortened for brevity). Examples are given for both a flat parse and a compositional parse. Our models are trained to generate the annotation sequences.}
\vspace{-0.5em}
\begin{tabular}{p{1.8cm}p{14.2cm}}
\toprule
\textbf{Complexity} & \textbf{NLU Annotation} \\
\midrule
Flat & play Jacques station \\
& \texttt{[IN:PLAY\_MUSIC [SL:PLAYLIST} Jacques \texttt{ ] [SL:TYPE} station \texttt{] ]} \\
\midrule
Compositional & driving directions to the Eagles game \\
& \texttt{[IN:DIRECTION [SL:DESTINATION [IN:EVENT [SL:NAME} Eagles \texttt{] [SL:CAT} game \texttt{ ] ] ] ]} \\
\bottomrule
\end{tabular}
\vspace{-1em}
\label{tab:nlu-annotations}
\end{table*}

\begin{figure}[t]
  \centering
  \includegraphics[width=\columnwidth]{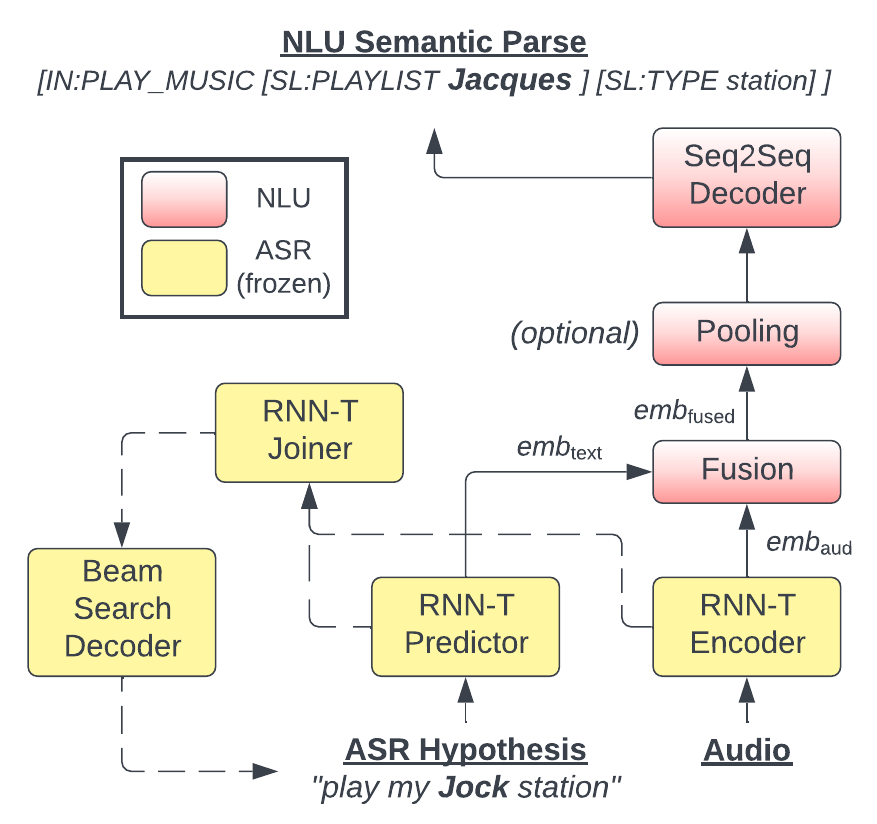}
  \vspace{-2em}
  \caption{System diagram for deliberation-based end-to-end SLU. ASR blocks are pre-trained with RNN-T loss on paired audio data and remain frozen during NLU finetuning. Dashed arrows represent the ASR pass to obtain the initial hypothesis.}
  \vspace{-1em}
  \label{fig:deliberation_diagram}
\end{figure}

Our proposed deliberation approach to E2E SLU is depicted in Figure~\ref{fig:deliberation_diagram}, where a separately trained ASR model is used to obtain the initial hypothesis given some audio input. The ASR model is assumed to be a variant of Recurrent Neural Network Transducer (RNN-T)~\cite{Graves12transduction}, the most popular streaming E2E ASR solution currently. Note that our approach can be easily modified to support other E2E ASR architectures. An important design choice in our system is that the ASR component remains \textbf{frozen}. This guarantees that the system can still generate transcripts at the same level of accuracy and latency, thus ASR remains general-purpose and can be shared with other tasks. This is especially important for on-device processing since we can reuse ASR parameters for downstream tasks, thus reducing the memory and computation requirement.

The initial ASR hypothesis is fed to the RNN-T predictor (analogous to a language model) to obtain the text embedding $emb_{\text{text}} \in \mathbb{R}^{T \times D}$, where $T$ is the text embedding length and $D$ is the embedding dimension. Similarly, the audio is fed to the RNN-T encoder (analogous to an acoustic model) to obtain the audio embedding $emb_{\text{aud}} \in \mathbb{R}^{A \times D}$, where $A$ is the audio embedding length and $D$ is the embedding dimension. For simplicity, we assume that the text and audio embeddings have the same dimension; if not the case, we can achieve this with a simple projection layer. In practice, $emb_{\text{text}}$ and $emb_{\text{aud}}$ are cached during ASR decoding and do not need to be re-computed.

The \textbf{\emph{Fusion}} module then combines $emb_{\text{text}}$ and $emb_{\text{aud}}$ into a single embedding $emb_{\text{fused}} \in \mathbb{R}^{T \times D}$ as follows:
\begin{align}
    emb_{\text{attn}}, \text{\underline{\hspace{0.7em}}} &= \texttt{MHA}(emb_{\text{text}}, emb_{\text{aud}}, emb_{\text{aud}}) \\
    emb_{\text{stack}} &= \texttt{Stack}(emb_{\text{text}}, emb_{\text{attn}}) \\
    emb_{\text{fused}} &= \texttt{Linear}(emb_{\text{stack}})
\end{align}

Here $emb_{\text{attn}} \in \mathbb{R}^{T \times D}$ is the output of multi-head attention (MHA)~\cite{Vaswani2017} with $emb_{\text{text}}$ as query and $emb_{\text{aud}}$ as key and value. $emb_{\text{attn}}$ can be viewed as a semantically-induced version of the audio embeddings. $emb_{\text{stack}} \in \mathbb{R}^{T \times 2D}$ is obtained by concatenating $emb_{\text{text}}$ and $emb_{\text{attn}}$ in the feature dimension, and $emb_{\text{fused}} \in \mathbb{R}^{T \times D}$ is the result of projecting $emb_{\text{stack}}$ down to the original embedding dimension $D$. This fusion method empirically performed the best compared to various approaches we experimented with, including embedding concatenation in time and cross-attention interface~\cite{raju2021end}. $emb_{\text{fused}}$ effectively combines text and audio signal into a single embedding while keeping the sequence length compact (typically $T \ll A$).

$emb_{\text{fused}}$ is further encoded by a \textbf{\emph{Pooling}} module consisting of $N$ transformer layers, before being fed to a transformer \textbf{\emph{Decoder}} to generate the target semantic parse. Critically, to avoid requiring our model to regenerate every token, we introduce a \textbf{\emph{pointer-generator head}} \cite{see2017get, decoupled}. Given the decoder state for the current timestep $d_t$ and encoder outputs $e$, the copy head computes (1) probability of generating a new token $g_t$ (2) probability of copying a token $c_t$ by relying on the initial ASR hypothesis $h$ and the attention vector $\omega_t$, and (3) a mixing probability $P_\text{copy}$ to mix the copy and generation distribution to obtain the final output token distribution $o_t$:
\begin{align}
    g_t &= \texttt{Softmax}(\texttt{Linear}(d_t)) \\ 
    \gamma_t, \omega_t &=  \texttt{MHA}(d_t, e, e) \\
    c_t &= \texttt{Scatter}(h, \omega_t) \\
    P_\text{copy} &=  \sigma(\texttt{Linear}([d_t, \gamma_t])) \\
    o_t &= \left(1-P_\text{copy}\right) \cdot g_t + P_\text{copy} \cdot c_t
\end{align}

In training, the loss is backpropagated from the target down to the NLU components, while the ASR blocks remain frozen. Unlike traditional NLU models, our proposed approach is theoretically able to correct ASR errors in the NLU pass, as demonstrated in Figure~\ref{fig:deliberation_diagram} where ``Jock" is corrected to ``Jacques".

\vspace{-0.5em}
\section{Experimental Setup}
\label{sec:experimental_setup}

\subsection{SLU Dataset and Evaluation Procedure}

To evaluate the scale and robustness of our approach, we evaluate our models on the spoken version of the popular open-source NLU benchmark, Task-Oriented Semantic Parsing (TOPv2) \cite{chen-2020-topv2}. This dataset consists of eight domains across: alarm, messaging, music, navigation, timer, weather, reminder, and event, with both flat and compositional parse structures as shown in Table~\ref{tab:nlu-annotations}. The dataset is divided into three splits: \texttt{train} for model training (125k utterances), \texttt{valid} for hyperparameter tuning (17k utterances), and \texttt{test} for evaluation (39k utterances). For the spoken version of TOPv2, referred to as Spoken Task-Oriented Parsing (STOP), we collected speech data through Amazon Mechanical Turk (MTurk) where each utterance was spoken by up to two speakers\cite{stop2022}. The final corpus consists of 885 speakers and 218 hours of speech.

We evaluate SLU models with \textbf{Exact Match (EM)}~\cite{chen-2020-topv2}, which measures the accuracy of the hypothesis against the reference annotation with an exact string match (punctuation and casing are ignored). To be counted as correct, both the parse structure and the slot content transcription must be correct. For example, given the reference ``\texttt{[IN:PLAY\_MUSIC [SL:PLAYLIST} \textbf{Jacques} \texttt{][SL:TYPE} station \texttt{]]}", the hypothesis ``\texttt{[IN:PLAY\_MUSIC [SL:PLAYLIST} \textbf{Jock} \texttt{][SL:TYPE} station \texttt{]]}" is considered incorrect because the slot content ``Jacques" is mistranscribed as ``Jock".

\vspace{-0.5em}
\subsection{SLU Systems}
\label{ssec:slu}

As we target on-device processing in this work, let us assume two hypothetical devices with different computation budgets for ASR: \textbf{D1} (25M parameters) and \textbf{D2} (10M parameters). In both devices, NLU has a fixed budget of \textbf{5M} parameters. All models will adhere to these budgets\footnote{We experimented with many hypothetical budgets and the trends were similar; we consider only these scenarios in the paper for brevity.}.

\vspace{-0.5em}
\subsubsection{ASR Models}
\label{sssec:asr}

Both ASR models used in this work are variants of RNN-T which employ a 1-layer LSTM predictor, a streamable low-latency Conformer encoder~\cite{Shi2022conformer} with varying number of layers (13 for D1, 3 for D2), and a 1-layer fully connected joiner network. The encoder has a stride of four, 40ms lookahead, and 120ms segment size. The target units are 4095 unigram WordPieces~\cite{Kudo2018SubWord} built with SentencePiece~\cite{kudo-richardson-2018-sentencepiece}. The model is trained on 145k hours of in-house speech data using Alignment Restricted RNN-T loss~\cite{Mahadeokar2021AR-RNNT} and SpecAugment~\cite{park2019specaugment}, following the recipe in~\cite{le21_interspeech}. Note that STOP is \textbf{not} included in the ASR training data. The total size and Word Error Rate (WER) on STOP \texttt{test} for these models are as follows: \textbf{D1} -- 25.0M parameters, 3.54\% WER; \textbf{D2} -- 9.8M parameters, 6.54\% WER.

The role of these ASR models is to provide the first-pass hypothesis to the pipeline NLU model, as well as the text and audio embeddings (via the RNN-T predictor and encoder) to the deliberation NLU model.

\vspace{-0.5em}
\subsubsection{Pipeline NLU Baseline}
\label{sssec:nlu}

Our pipeline text-based NLU baseline adopts the encoder-decoder non-autoregressive architecture~\cite{nar_semantic_parsing} with span pointer network~\cite{spanpointernetwork}. This setup demonstrated superior results over non-autoregressive and autoregressive alternatives at smaller sizes on TOPv2~\cite{spanpointernetwork}. The architecture consists of a transformer encoder based on the RoBERTa architecture~\cite{roberta}; however we randomly initialize the model instead of bootstrapping it with RoBERTa layers, as using only one RoBERTa layer will already exceed the NLU budget. We set the transformer embedding dimension to 256, the number of attention heads to 8, and the number of encoder layers to 3. We follow \cite{nar_semantic_parsing, spanpointernetwork} and have a single layer of span pointer transformer decoder. All models adopt the same unigram WordPiece tokenization used by ASR and are trained on TOPv2 \texttt{train}, with training hyperparameters optimized on TOPv2 \texttt{valid}.

The final model contains \textbf{4.8M} parameters and achieves an EM of \textbf{83.47\%} on TOPv2 \texttt{test}. This result is competitive given the model's size; a 134M-parameter RoBERTa-bootstrapped version achieved 3.78\% higher EM accuracy~\cite{spanpointernetwork}.

\vspace{-0.5em}
\subsubsection{Deliberation NLU Models}
\label{sssec:deliberation_models}

Our deliberation model uses the frozen first-pass RNN-T for extracting text and audio embeddings as described in Section~\ref{sec:method}. Specifically, we use the second-to-last hidden layer (before the final linear layer) of both the predictor and encoder, which gives an initial feature dimension of 256. The \emph{Fusion} module's MHA component employs 8 attention heads. The \emph{Pooling} module consists of 2 transformer encoder layers with 8 attention heads and an embedding dimension of 224. Finally, the \emph{Decoder} module contains a single transformer decoder layer with 2 attention heads and a pointer-generator with 1 attention head. The targets are the TOPv2 annotation sequences tokenized using ASR's SentencePiece model and 586 ontology tokens, resulting in 4681 output units. The model contains \textbf{4.7M} parameters.

In order to analyze the importance of multimodality, we also train text-only and audio-only versions of our deliberation models. For the text-only version, we replace $emb_{\text{fused}}$ with $emb_{\text{text}}$ and increase the \emph{Pooling} transformer's embedding dimension to 240 since the model no longer has a \emph{Fusion} module, resulting in \textbf{4.9M} parameters. For the audio-only version, we replace $emb_{\text{fused}}$ with $emb_{\text{aud}}$ and further increase the \emph{Pooling} transformer's embedding dimension to 256 since both the \emph{Fusion} module and the pointer generator head are no longer applicable, resulting in \textbf{5.0M} parameters.

We train the models using Cross Entropy loss with 0.1 label smoothing and SpecAugment applied on the raw audio features (before going through the encoder), with training hyperparameters optimized on STOP \texttt{valid}.

Our deliberation models can use either the first-pass ASR hypothesis or the reference as the input text for training. We also explore duplicating every utterance with ASR error and use both the hypothesis and reference for training. This \textbf{union} strategy teaches the model how to handle ASR errors while ensuring full coverage of the TOPv2 training text. Note that the pipeline NLU baseline only allows training on the reference since the model generates spans (indices into the input) instead of tokens.

\begin{table}[t]
\centering
\caption{Pipeline vs. deliberation SLU results on STOP \texttt{test}. \textbf{Text} indicates whether the model is trained using the first-pass ASR hypothesis (\emph{Hyp}), reference text (\emph{Ref}), or the union of both (\emph{Union}). All NLU models have $\sim$5M parameters.}
\vspace{-0.5em}
\begin{tabular}{c c c c c}
\toprule
\textbf{ASR} & \textbf{NLU} & \textbf{Text} & \textbf{Modality} & \textbf{EM} \\
\midrule
\midrule
\multirow{8}{*}{D1} & \emph{Pipeline} & \emph{Ref} & \emph{Text} & \emph{73.22} \\
\cline{2-5}
& \multirow{7}{*}{\textbf{Deliberation}} & - & Audio & 69.99 \\
\cline{3-5}
& & \multirow{2}{*}{Hyp} & Text & 72.55 \\
& & & Fusion & 72.83 \\
\cline{3-5}
& & \multirow{2}{*}{Ref} & Text & 72.96 \\
& & & Fusion & 73.20 \\
\cline{3-5}
& & \multirow{2}{*}{\textbf{Union}} & Text & 73.22 \\
& & & \textbf{Fusion} & \textbf{73.87} \\
\midrule
\midrule
\multirow{8}{*}{D2} & \emph{Pipeline} & \emph{Ref} & \emph{Text} & \emph{67.30} \\
\cline{2-5}
& \multirow{7}{*}{\textbf{Deliberation}} & - & Audio & 65.47 \\
\cline{3-5}
& & \multirow{2}{*}{Hyp} & Text & 66.98 \\
& & & Fusion & 66.92 \\
\cline{3-5}
& & \multirow{2}{*}{Ref} & Text & 67.10 \\
& & & Fusion & 66.91 \\
\cline{3-5}
& & \multirow{2}{*}{\textbf{Union}} & Text & 67.84 \\
& & & \textbf{Fusion} & \textbf{67.90} \\
\bottomrule
\bottomrule
\end{tabular}
\vspace{-1em}
\label{tab:delib-results}
\end{table}

\vspace{-0.5em}
\section{Results and Discussion}
\label{sec:results}

\subsection{Pipeline vs. Deliberation}
\label{ssec:deliberation_results}

As seen in Table~\ref{tab:delib-results}, our proposed deliberation approach outperforms the pipeline baseline on both devices, with the absolute EM improvement ranging from \textbf{0.60\%} to \textbf{0.65\%}. The EM gap between pipeline and deliberation is larger for D1, which suggests that our approach is more beneficial when the base ASR model is stronger. Both pipeline and deliberation methods are heavily dependent on ASR quality; the more accurate ASR is, the better NLU performs and vice versa. The overall results confirm that our deliberation-based approach is especially suitable for when computational resources are limited, such as on-device. For server-side processing, the pipeline approach benefits significantly from text-based pre-training (e.g., from RoBERTa); the same benefit may not extend to the multimodal deliberation model. We will explore pre-training for deliberation systems in future work.

The \emph{Union} training strategy consistently outperforms either \emph{Hyp} or \emph{Ref} in isolation. This suggests that it is important to teach the model how to correct ASR errors while exposing the model to as much text data as possible.

Among the different modalities, text-only performs better than audio-only, pointing to the importance of ASR. Text and audio fusion performs better than text-only for D1, but not for D2. The advantage of fusion becomes smaller as the base ASR model gets weaker, with the EM improvement over text-only baselines reducing from 0.65\% for D1 to just 0.06\% for D2. This means that as the quality of the audio embeddings decreases, they no longer offer complementary benefits over the text embeddings. We will explore methods to improve the audio embedding quality of small ASR models in future work.

\vspace{-0.5em}
\subsection{Robustness to ASR Errors}
\label{ssec:robustness}

\begin{table}[t]
\centering
\caption{EM results on STOP \texttt{test} for \emph{D1}, broken down by whether or not there is an ASR error.}
\vspace{-0.5em}
\begin{tabular}{c c c c}
\toprule
\multirow{2}{*}{\textbf{NLU}} & \multirow{2}{*}{\textbf{Modality}} & \multicolumn{2}{c}{\textbf{ASR Error}} \\
& & \textbf{\emph{No (61k utts)}} & \textbf{\emph{Yes (14k utts)}} \\
\midrule
\emph{Pipeline} & \emph{Text} & \emph{84.47} & \emph{24.49} \\
\midrule
\multirow{3}{*}{Deliberation} & Text & 83.12 & 30.32 \\
& Audio & 77.32 & 38.10 \\
& Fusion & 83.68 & 31.42 \\
\bottomrule
\end{tabular}
\vspace{-1em}
\label{tab:robustness-results}
\end{table}

\begin{table}[t]
\centering
\caption{Example utterances where the \emph{Fusion} deliberation model for \emph{D1} was able to recover from ASR errors.}
\vspace{-0.5em}
\begin{tabular}{p{0.445\columnwidth} p{0.445\columnwidth}}
\toprule
\textbf{Reference} & \textbf{ASR Hypothesis} \\
\midrule
...a \textbf{\emph{land or a sea breeze}}... & ...a \textbf{\emph{lander sebree's}}... \\
...for \textbf{\emph{sao paulo}}... & ...for \textbf{\emph{south palo}}... \\
...forecast for \textbf{\emph{newcastle}} & ...forecast for \textbf{\emph{new castle}} \\
\textbf{\emph{get rid}} of the reminder... & \textbf{\emph{garrett}} of the reminder... \\
\textbf{\emph{laughs}} for john's message & \textbf{\emph{laps}} for john's message \\
insert \textbf{\emph{heart}} eyes to... & insert \textbf{\emph{her}} eyes to... \\
\bottomrule
\end{tabular}
\vspace{-1.5em}
\label{tab:robustness-examples}
\end{table}

One major appeal of deliberation models is their ability to rewrite the initial hypothesis in the NLU pass, which makes the system more robust to ASR errors. We quantify ASR robustness in Table~\ref{tab:robustness-results} by splitting the EM results into two buckets, one containing utterances with no ASR error and one with at least one error. As can be seen, all deliberation variants significantly outperform the pipeline NLU baseline on utterances with ASR errors, at the cost of underperforming on correctly transcribed utterances. The audio-only deliberation model performs the best on the error bucket since it bypasses the initial hypothesis completely. The text-only model, by contrast, performs better on the error-free bucket but worse on the error bucket since it does not have access to the audio. The fusion of text and audio features provides the best overall balance between the two buckets. Future work will focus on improving the numbers for both buckets, possibly by unifying deliberation-based ASR and span pointer-based NLU in a single framework.

Table~\ref{tab:robustness-examples} shows some examples where the fusion deliberation model was able to recover from ASR errors whereas both the pipeline baseline and the text-only deliberation model were not. In most cases, recovery was possible due to the correction of small ASR errors that affect the semantic of the entire utterance.

\vspace{-0.5em}
\subsection{Scaling E2E SLU with Synthetic Speech}
\label{ssec:tts}

One of the biggest limitations of E2E SLU compared to the traditional text-only NLU is the former's reliance on audio training data, as collecting natural speech is often significantly more expensive than collecting text data. Thus, it is usually more difficult and time-consuming to scale E2E SLU to new domains. With its many advances in recent years, text-to-speech (TTS) offers a potential solution to the scalability of E2E SLU by replacing natural speech with synthetic speech for model training. We explore this possibility with deliberation models by replacing the STOP \texttt{train}'s natural speech with synthetic speech generated by our in-house TTS engine. Each utterance is synthesized using a randomly chosen voice among eight available voice profiles, together with a randomly adjusted pitch and speaking rate to further increase the speaker diversity.

\begin{table}[t]
\centering
\caption{Deliberation SLU results on STOP \texttt{test} when using natural vs. synthetic (TTS-generated) speech for training.}
\vspace{-0.5em}
\begin{tabular}{c c c c c}
\toprule
\textbf{ASR} & \textbf{Modality} & \textbf{EM (Natural)} & \textbf{EM (TTS)} \\
\midrule
\multirow{2}{*}{D1} & Audio & 69.99 & 67.10 (-2.89) \\
& Fusion & 73.87 & 71.96 (-1.91) \\
\midrule
\multirow{2}{*}{D2} & Audio & 65.47 & 60.15 (-5.32) \\
& Fusion & 67.90 & 66.34 (-1.56) \\
\bottomrule
\end{tabular}
\vspace{-1.5em}
\label{tab:tts-results}
\end{table}

The results for training with synthetic speech are summarized in Table~\ref{tab:tts-results}. Audio-only systems appear particularly sensitive to the mismatch in audio training data, and weaker RNN-T encoders cannot generalize from TTS data as effectively (e.g., the EM degradation is 2.89\% for D1 vs. 5.32\% for D2). Text and audio fusion is able to significantly reduce the degradation due to TTS training; however, the final EM still lags behind the text-only EM from Table~\ref{tab:delib-results}. These results suggest that while deliberation makes TTS-based training more viable due to the combination of text and audio features, more work is required to make E2E SLU scalable using purely TTS-generated speech. We will explore this direction further in future work.

\section{Conclusion and Future Work}
\label{sec:conclusion}

In this paper, we proposed a novel deliberation-based approach to E2E SLU which fuses text embeddings (derived from the first-pass ASR hypothesis) with audio embeddings to generate the target semantic parse. We demonstrate that our proposed approach is able to consistently outperform strong pipeline baselines across two on-device operating points, is more robust to ASR errors, and makes TTS-only training more viable. For future work, we plan to explore pre-training strategies for deliberation, improve the audio embedding quality of small ASR models, unify our approach with deliberation-based ASR and span pointer-based NLU, and further push the results on pure TTS training sets to make E2E SLU more scalable.

\newpage
\bibliographystyle{IEEEtran}

\bibliography{mybib}

\begin{thebibliography}{10}
\providecommand{\url}[1]{#1}
\csname url@samestyle\endcsname
\providecommand{\newblock}{\relax}
\providecommand{\bibinfo}[2]{#2}
\providecommand{\BIBentrySTDinterwordspacing}{\spaceskip=0pt\relax}
\providecommand{\BIBentryALTinterwordstretchfactor}{4}
\providecommand{\BIBentryALTinterwordspacing}{\spaceskip=\fontdimen2\font plus
\BIBentryALTinterwordstretchfactor\fontdimen3\font minus
  \fontdimen4\font\relax}
\providecommand{\BIBforeignlanguage}[2]{{%
\expandafter\ifx\csname l@#1\endcsname\relax
\typeout{** WARNING: IEEEtran.bst: No hyphenation pattern has been}%
\typeout{** loaded for the language `#1'. Using the pattern for}%
\typeout{** the default language instead.}%
\else
\language=\csname l@#1\endcsname
\fi
#2}}
\providecommand{\BIBdecl}{\relax}
\BIBdecl

\bibitem{Haghani18SLU}
P.~Haghani, A.~Narayanan, M.~A.~U. Bacchiani, G.~Chuang, N.~Gaur, P.~J.~M.
  Mengibar, D.~Qu, R.~Prabhavalkar, and A.~Waters, ``From audio to semantics:
  Approaches to end-to-end spoken language understanding,'' in \emph{Proc.
  SLT}, 2018.

\bibitem{serdyuk2018endtoend}
D.~Serdyuk, Y.~Wang, C.~Fuegen, A.~Kumar, B.~Liu, and Y.~Bengio, ``Towards
  end-to-end spoken language understanding,'' 2018.

\bibitem{Potdar2021ASE}
N.~Potdar, A.~R. Avila, C.~Xing, D.~Wang, Y.~Cao, and X.~Chen, ``A streaming
  end-to-end framework for spoken language understanding,'' in \emph{IJCAI},
  2021.

\bibitem{Radfar2021FANSFA}
M.~Radfar, A.~Mouchtaris, S.~Kunzmann, and A.~Rastrow, ``Fans: Fusing asr and
  nlu for on-device slu,'' in \emph{Interspeech}, 2021.

\bibitem{Wang2021Speech2SlotAE}
P.~Wang, X.~Ye, X.~Zhou, J.~Xie, and H.~Wang, ``Speech2slot: An end-to-end
  knowledge-based slot filling from speech,'' \emph{ArXiv}, vol.
  abs/2105.04719, 2021.

\bibitem{2020NeuralInterface}
M.~Rao, A.~Raju, P.~Dheram, B.~Bui, and A.~Rastrow, ``Speech to semantics:
  Improve asr and nlu jointly via all-neural interfaces,'' in \emph{Proc.
  INTERSPEECH}, 2020.

\bibitem{raju2021end}
A.~Raju, G.~Tiwari, M.~Rao, P.~Dheram, B.~Anderson, Z.~Zhang, B.~Bui, and
  A.~Rastrow, ``End-to-end spoken language understanding using rnn-transducer
  asr,'' \emph{arXiv preprint arXiv:2106.15919}, 2021.

\bibitem{decoupled}
A.~Aghajanyan, J.~Maillard, A.~Shrivastava, K.~Diedrick, M.~Haeger, H.~Li,
  Y.~Mehdad, V.~Stoyanov, A.~Kumar, M.~Lewis, and S.~Gupta, ``{Conversational
  Semantic Parsing},'' in \emph{Proceedings of the Conference on Empirical
  Methods in Natural Language Processing and the International Joint Conference
  on Natural Language Processing (EMNLP-IJCNLP)}, 2020.

\bibitem{gupta2018semantic}
S.~Gupta, R.~Shah, M.~Mohit, A.~Kumar, and M.~Lewis, ``{Semantic Parsing for
  Task Oriented Dialog using Hierarchical Representations},'' in
  \emph{Proceedings of the Conference on Empirical Methods in Natural Language
  Processing (EMNLP)}, 2018.

\bibitem{rongali2020don}
S.~Rongali, L.~Soldaini, E.~Monti, and W.~Hamza, ``{Don't Parse, Generate! A
  Sequence to Sequence Architecture for Task-Oriented Semantic Parsing},'' in
  \emph{Proceedings of the Web Conference (WWW)}, 2020.

\bibitem{Hu20deliberation}
``Deliberation model based two-pass end-to-end speech recognition,'' in
  \emph{Proc. ICASSP}, K.~Hu, R.~Prabhavalkar, R.~Pang, and T.~Sainath, Eds.,
  2020.

\bibitem{Graves12transduction}
A.~Graves, ``Sequence transduction with recurrent neural networks,'' in
  \emph{ICML Representation Learning Workshop}, 2012.

\bibitem{Vaswani2017}
A.~Vaswani, N.~Shazeer, N.~Parmar, J.~Uszkoreit, L.~Jones, A.~N. Gomez, L.~u.
  Kaiser, and I.~Polosukhin, ``Attention is all you need,'' in \emph{Proc.
  NIPS}, 2017.

\bibitem{see2017get}
A.~See, P.~J. Liu, and C.~D. Manning, ``Get to the point: Summarization with
  pointer-generator networks,'' in \emph{Proceedings of the 55th Annual Meeting
  of the Association for Computational Linguistics (Volume 1: Long Papers)},
  2017.

\bibitem{chen-2020-topv2}
X.~Chen, A.~Ghoshal, Y.~Mehdad, L.~Zettlemoyer, and S.~Gupta, ``{Low-Resource
  Domain Adaptation for Compositional Task-Oriented Semantic Parsing},'' in
  \emph{Proceedings of the Conference on Empirical Methods in Natural Language
  Processing (EMNLP)}, 2020.

\bibitem{stop2022}
P.~Tomasello, A.~Shrivastava, D.~Lazar, P.-C. Hsu, D.~Le, A.~Sagar, A.~Elkahky,
  J.~Copet, W.-N. Hsu, Y.~Mordechay, R.~Algayres, T.~A. Nguyen, E.~Dupoux,
  L.~Zettlemoyer, and A.~Mohamed, ``{STOP: A dataset for Spoken Task Oriented
  Semantic Parsing},'' in \emph{CoRR}.

\bibitem{Shi2022conformer}
Y.~Shi, C.~Wu, D.~Wang, A.~Xiao, J.~Mahadeokar, X.~Zhang, C.~Liu, K.~Li,
  Y.~Shangguan, V.~Nagaraja \emph{et~al.}, ``Streaming transformer transducer
  based speech recognition using non-causal convolution,'' \emph{Proc. ICASSP},
  2022.

\bibitem{Kudo2018SubWord}
T.~Kudo, ``{Subword Regularization: Improving Neural Network Translation Models
  with Multiple Subword Candidates},'' in \emph{Proc. ACL}, 2018.

\bibitem{kudo-richardson-2018-sentencepiece}
T.~Kudo and J.~Richardson, ``{{S}entence{P}iece: A simple and language
  independent subword tokenizer and detokenizer for Neural Text Processing},''
  in \emph{Proc. EMNLP: System Demonstrations}, 2018.

\bibitem{Mahadeokar2021AR-RNNT}
J.~Mahadeokar, Y.~Shangguan, D.~Le, G.~Keren, H.~Su, T.~Le, C.~Yeh, C.~Fuegen,
  and M.~L. Seltzer, ``{Alignment Restricted Streaming Recurrent Neural Network
  Transducer},'' in \emph{Proc. SLT}, 2021.

\bibitem{park2019specaugment}
D.~Park, W.~Chan, Y.~Zhang, C.~Chiu, B.~Zoph, E.~Cubuk, and Q.~Le,
  ``{SpecAugment: A simple data augmentation method for automatic speech
  recognition},'' in \emph{Proc. INTERSPEECH}, 2019.

\bibitem{le21_interspeech}
D.~Le, M.~Jain, G.~Keren, S.~Kim, Y.~Shi, J.~Mahadeokar, J.~Chan, Y.~Shangguan,
  C.~Fuegen, O.~Kalinli, Y.~Saraf, and M.~L. Seltzer, ``{Contextualized
  Streaming End-to-End Speech Recognition with Trie-Based Deep Biasing and
  Shallow Fusion},'' in \emph{Proc. Interspeech}, 2021, pp. 1772--1776.

\bibitem{nar_semantic_parsing}
A.~Babu, A.~Shrivastava, A.~Aghajanyan, A.~Aly, A.~Fan, and M.~Ghazvininej,
  ``{Non-Autoregressive Semantic Parsing for Compositional Task-Oriented
  Dialog},'' in \emph{Proceedings of the Annual Conference of the North
  American Chapter of the Association for Computational Linguistics (NAACL)},
  2021.

\bibitem{spanpointernetwork}
A.~Shrivastava, P.~Chuang, A.~Babu, S.~Desai, A.~Arora, A.~Zotov, and A.~Aly,
  ``{Span Pointer Networks for Non-Autoregressive Task-Oriented Semantic
  Parsing},'' in \emph{Proceedings of the Findings of the Conference on
  Empirical Methods in Natural Language Processing (EMNLP)}, 2021.

\bibitem{roberta}
Y.~Liu, M.~Ott, N.~Goyal, J.~Du, M.~Joshi, D.~Chen, O.~Levy, M.~Lewis,
  L.~Zettlemoyer, and V.~Stoyanov, ``{RoBERTa: A Robustly Optimized BERT
  Pretraining Approach},'' \emph{arXiv preprint arXiv:1907.11692}, 2019.

\end{thebibliography}

\end{document}